\def\year{2020}\relax
\documentclass[letterpaper]{article} 
\usepackage{aaai20}  
\usepackage{times}  
\usepackage{helvet} 
\usepackage{courier}  
\usepackage[hyphens]{url}  
\usepackage{graphicx} 
\usepackage{graphicx}  
\usepackage{algorithm}
\usepackage{algorithmic}
\usepackage{amsfonts}   

\usepackage[utf8]{inputenc} 
\usepackage[T1]{fontenc}    
\usepackage{booktabs}       
\usepackage{amsfonts}       
\usepackage{nicefrac}       
\usepackage{microtype}      
\usepackage{algorithm}
\usepackage{algorithmic}
\usepackage{enumitem}
\usepackage{textcomp}
\usepackage{color}
\usepackage{makecell}

\pdfoutput=1
\title{Machine learning  discrimination of Parkinson's Disease stages from walker-mounted sensors data}

\author{Nabeel Seedat \\ University of the Witwatersrand, Johannesburg \\ seedatnabeel@gmail.com
\And
Vered Aharonson \\ University of the Witwatersrand, Johannesburg \\ vered.aharonson@wits.ac.za}
\begin{document}

\maketitle

\begin{abstract}
Clinical methods that assess gait in Parkinson's Disease (PD) are mostly qualitative. Quantitative methods necessitate costly instrumentation or cumbersome wearable devices, which limits their usability. Only few of these methods can discriminate different stages in PD progression.  This study applies machine learning methods to discriminate six stages of PD. The data was acquired by low cost walker-mounted sensors in an experiment at a movement disorders clinic and the PD stages were clinically labeled. A large set of features, some unique to this study are extracted and three feature selection methods are compared using a multi-class Random Forest (RF) classifier.  The feature subset selected by the Analysis of Variance (ANOVA) method  provided performance similar to the full feature set:  93\% accuracy and had significantly shorter computation time. Compared to PCA, this method also enabled clinical interpretability of the selected features, an essential attribute to healthcare applications.  All selected-feature sets are dominated by information theoretic features and statistical features and offer insights into the characteristics of gait deterioration in PD. The results indicate a feasibility of machine learning to accurately classify PD severity stages from kinematic signals acquired by low-cost, walker-mounted sensors and implies a potential to aid medical practitioners in the quantitative assessment of PD progression. The study presents a solution to the small and noisy data problem, which is common in most sensor-based healthcare assessments. 
\end{abstract}

\section{Introduction}

Automated disease diagnosis has the potential to reduce labor and cost in healthcare, as well as 
offer an augmented accuracy which may improve treatment efficacy. Machine learning methods that 
could provide automated disease assessment have been extensively studied on medical imaging and 
physiological signals datasets. 

These studies reported accurate detection of diseases and dysfunctions such as  cardiovascular  disease \cite{poplin2018prediction},  kidney dysfunction \cite{tomavsev2019clinically},  brain tumors \cite{Havaei2017BrainTS} and more.  When  deep learning methods are employed, very large datasets were needed, however, to provide accurate detection or 
classification of diseases. Such datasets are not available in many healthcare contexts.

For patients suffering from Parkinson's disease; a debilitating neuro-degenerative disease with 
increasing prevalence; datasets are typically scarce and small. Parkinson's disease severity determination is predominantly based on observational or reported symptoms which are challenging to reliably acquire in an automated manner. Healthcare professionals and patients could greatly benefit from automated disease assessment tools  since observation- or reporting-based data are qualitative and subjective, and hence are prone to bias and inaccuracy, which may lead to incorrect treatment. 

Attempts to automate Parkinson's disease assessment typically use sensors that capture patients' motor data \cite{Jarchi}. Although sensors have the potential to efficiently record large quantities of quantitative and objective data, practical considerations often limit their usage in clinical settings: Many sensors and devices are cumbersome and uncomfortable, and/or are not suitable 
for severe stages of Parkinson's disease, when patients cannot walk without support \cite{Aharonson2018,Bryant2014}. The small data size may also be the reason that the goal in most previous works was to discriminate between patients and controls, without attempting the multi-class discrimination of disease stages. An accurate and timely discrimination of  disease progression stages is very important for efficient treatment and patient care.
\begin{figure*}[t]
 \centering
  \includegraphics[width=\linewidth]{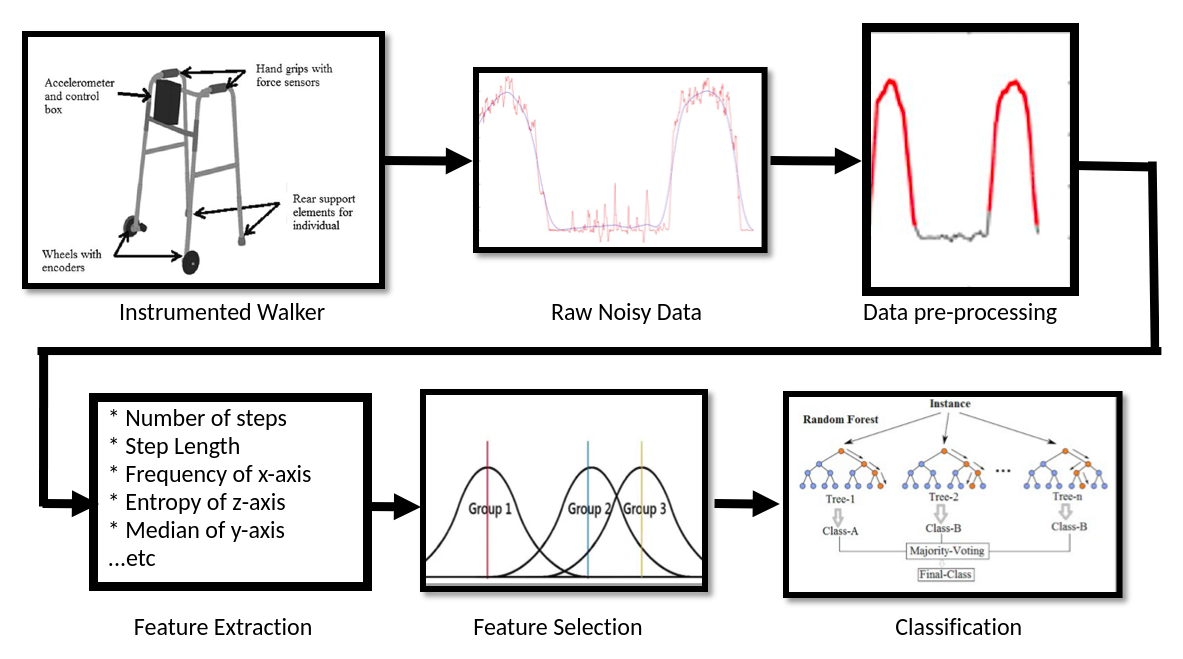}
  \caption{System Pipeline outlining the six stages of processing from raw data acquired from the instrumented walker till the classification stage of PD stage.}
  \label{pipeline}
\end{figure*}

 This paper makes the following contributions: 
\begin{itemize}
    \item A novel sensor-based system that offers a solution to the aforementioned limitations. This 
system collects adequate data on Parkinson's disease symptoms and uses machine learning for quantitative multi-class discrimination of Parkinson's disease stages. 
   \item A machine learning method that can provide in a translucent way the most relevant features 
that discriminate between Parkinson's disease stages. This method enables non-technological 
healthcare professionals to interpret the system's output and employ their interpretation in 
patient-care planning
	\item Solutions for the application of machine learning on small-scale datasets of noisy sensors' 
signals through an integration of signal processing techniques. 
\end{itemize}

\section{Background and Related Work}

\subsection{Parkinson's Disease}
Parkinson's disease (PD) affects more than 10 million individuals globally \cite{Lau2006}. This debilitating neuro-degenerative  disease manifests in both physical and mental symptoms \cite{Perumal2016,Sayeed2015}. Many of the physical symptoms pertain to gait; the manner in which a person walks \cite{Perumal2016,Sayeed2015}. 

Gait impairments reduces mobility, significantly  limits a person's functionality and independence and can cause instability and fall hazards.  Thus, the assessment of gait is an important component in discriminating the severity of Parkinson's disease \cite{Chen2013,dingwell2000nonlinear}. 

The Hoehn and Yahr scale is commonly employed by clinicians to reflect Parkinson's disease severity, providing severity stages between 1 and 5 \cite{Hoehn1998}.  As an extension to the integer severity scale, the modified Hoehn and Yahr scale has included a “mid-stage” of 2.5, in order to add granularity and reduce uncertainty in the discrimination \cite{Goetz2004}. 

The Hoehn and Yahr scale particularly pertains to gait and postural stability \cite{Hoehn1998}. While this scale is validated and has a standardized assessment procedure, the evaluation of the symptoms that provide its scores entails a subjective component \cite{Wang2014}. This affects the accuracy and objectivity of the diagnosis and hampers patient care. 

Moreover, the assessment is time-consuming and necessitates expert neurologists, which poses a burden on resource constrained healthcare settings and has limited accessibility to patients. 

The Timed-Up-and-Go (TUG) test, is sometimes used in addition to the Hoehn and Yahr assessment \cite{Aharonson2018,sprint,yahalom}. This test is simple, short and focuses only on gait: A patient sitting on a chair is requested to stand 
up, walk straight forward for 3m, turn around, walk back and sit on the chair, and the task completion time is logged. 

This method, however, only provides a single, rough measure that does not properly characterize the impairments entailed in the compound movement of the test.  Furthermore, this method still necessitates human observation and manual handling of a stopwatch or button, which limits its accuracy \cite{Aharonson2018,sprint,yahalom}.

\subsection{Quantitative Methods}
Automated assessments combining machine learning and sensor-based data acquisition are a potential 
solution to the aforementioned limitations of Parkinson's disease assessments. 

In previous studies, 
sensors such as accelerometers and gyroscopes, were of the strap-on or wearable type \cite{Mariani2012,Salarian2004,yahalom}. Moving with 
the patient's body, these sensors were able to capture kinematic data of the patient's gait. 

Although low-cost, most of these sensors are complex to strap on or wear and therefore are time-consuming to use for 
healthcare professionals, patients or caregivers. Additionally, some of these devices are cumbersome 
and uncomfortable to wear, thereby harming the user experience, especially for motor impaired 
persons \cite{Aharonson2018}. 

Moreover, in more severe cases of Parkinson's disease, patients require a walking aid such 
as a walker or cane, which further limits the usability and accuracy of wearables \cite{Bryant2014}. 

That being said, previous studies of sensors acquiring kinematic signals coupled with machine 
learning reported impressive algorithmic accuracy when discriminating two classes: patients 
from controls, or a when detecting a specific symptom.

Accelerometers placed on different parts of 
the body and a neural network classifier were able to discriminate between the presence or absence 
of dyskinesia in PD patients with an accuracy of 96.7\% \cite{keijsers2003automatic}. 

Smartphone accelerometer data and a random forest classifier yielded a discrimination of PD patients from controls with a sensitivity 
of 96.2\% and specificity of 96.9\% \cite{arora}. 

Infrared sensors used with a neural network classifier and a 
Support Vector Machine (SVM) discriminated PD patients from controls with an accuracy of 98\% 
and 100\%, respectively \cite{Tahir2012}. 

 To the best of our knowledge, no full sensor-based solution for sensor-based data acquisition overcoming the aforementioned usability limitations has been proposed, and no studies attempted a sensor-based classification of Hoehn and Yahr severity stages of PD.

\section{Our Approach: Machine learning to enable Parkinson's Disease stage discrimination from low-cost walker mounted sensor data}

We present a novel solution where machine learning is used for Parkinson's disease stage discrimination, as labelled by the modified Hoehn and Yahr severity scale.

The solution includes kinematic signals 
acquisition by walker-mounted sensors. The walker method provides a patient-centric approach that 
combines reliability and usability, even for severe stages of the disease. 

The solution includes techniques that can provide an accurate discrimination for small 
datasets, and for noisy signals, conditions that are typical to many healthcare datasets. 

Finally, our solution provides listing and ranking of explicable features in terms of their contribution to PD stage discrimination, which can readily be interpreted and used by clinicians for decision making in their patients assessment and treatment. 

 The analysis pipeline outlined in Figure \ref{pipeline} is implemented as follows:

\subsection{Dataset}
  The data used in this study was acquired by a low-cost aluminum walking frame retrofitted with low-cost and off-the-shelf
kinematic sensors.  The usage of a walking frame allows the subject's gait to be monitored unobtrusively, whilst
simultaneously supporting walking. 

The latter trait of supporting the subject is particularly important in advanced stages of the 
disease, when a walker is necessary for mobility. 

The sensors include a tri-axial accelerometer, distance encoders on the wheels, and force sensors on 
the handles that measure grip strength. 

The data acquired consists of the following seven signals: 

\begin{itemize}
    \item two force signals
    \item two distance encoder signals
    \item three acceleration signals (x,y,z axis)
\end{itemize}

Sixty-seven PD patients and nineteen age-matched healthy controls were subjected to the timed up and go (TUG) tests while using the walker. 

PD severity scores were assigned using the modified Hoehn and Yahr severity scores of 1, 2, 2.5, 3 and 4. Hoehn and Yahr scores of 5 were not included as these patients are bed ridden or confined to a wheelchair without assistance and are unable to perform the walking test \cite{Hoehn1998}. 

The Hoehn and Yahr stage was determined during a routine clinical assessment by a movement disorders neurologist.  

The study was conducted in a Movement Disorders Institute of a tertiary Medical Center. It was approved by the local institutional review board and all subjects signed an informed consent.

\subsection{Data pre-processing}
Pre-processing can enhance the reliability of feature extraction and is particularly important when 
 the data acquired is small in size. Small data size is common in experimental data acquisition where busy clinic conditions and patient recruitment challenges preclude large number of measurements. 
 
Pre-processing is necessary for most sensors' 
data, and particularly in the case of low-cost sensors, that often entail noisy, less accurate output. In our specific case of walker-mounted sensors, where the sensors do not move with the subject's body, a tailored pre-processing is needed in order to correctly detect footfalls – the basic elements of gait. 

The pre-processing of the walker's signals include noise and artifact removal, footfall detection, 
and segmentation of the walking phases into straight-line walking and turning phases: All 
the signals are denoised using a zero-phase lag moving average filter, thereby mitigating phase 
distortion. 

A Rodriquez Rotation matrix is employed to rotate the accelerometer signal and align it with the direction of subject movement. 

An adaptive Empirical Mode Decomposition (EMD) 
algorithm is then applied to the accelerometer signals in order to further denoise these signals for 
the footfall detection algorithm.

Our novel footfall detection algorithm is applied and  
 validated using video-camera manually-labeled data. Our algorithm yields an accuracy of 86\%, a 
significant improvement over a previous walker-based footfall detection algorithm from \cite{Ballesteros2017}.

\newpage
\subsection{Feature extraction}
The extracted feature set provides a quantitative representation of a subject's gait and are an amalgamation of novel features pertinent to the system used in this study and features widely used in previous studies. 

However, the derivation of the latter features need to be adjusted for the signals of the walker-mounted sensor system, where sensor type, quality and placement are different from previous studies. 

In total 211 features are extracted across four broad categories: spatio-temporal features, statistical features, frequency domain features and information theoretic features; 

\begin{enumerate}
  \item  Spatio-temporal features (which quantify the characteristics used by clinicians in their assessments of gait), such as step length, Timed-Up-and-Go (TUG) walk time, number of steps and turn time.
  \item Statistical Features, such as skewness, kurtosis, measures of central tendency, quartiles, mean and standard deviation of the signals' segments.  
\item Frequency Domain Features such as mean frequency, 3dB bandwidth,  99\% occupied bandwidth, 99\% occupied power, half power. 
\item  Information Theoretic Features such as cross-entropy, mutual information between axis, correlation between axes, walk ratio, walking intensity, harmonic ratio, zero cross rate.

\end{enumerate}

All accelerometer features are computed for the 3-axis of the accelerometer (the Cartesian x,y and z) 
as well as for their spherical co-ordinates: azimuth angle, elevation (polar) angle and radial distance.

\begin{table*}[t]
\centering
\caption{Accuracy and full feature set and the three selected subsets in the classification of Hoehn and Yahr PD stages  }
\label{sample-table}
\vskip 0.15in
\begin{sc}
\begin{tabular}{ccc}
\toprule

Test set \textbackslash Performance measure & \textbf{Accuracy (\%)} & \textbf{Mean Execution Time (s)}  \\
\midrule
All Features    & 94 $\pm$ 1.02 & 41.4\\
\textbf{ANOVA selected}    & \textbf{93} $\pm$ \textbf{1.3} & \textbf{15.2 }  \\
RF selected    & 90 $\pm$ 1.7 & 6.36    \\
PCA  & 95 $\pm$ 1.1 &  10.6  \\
\bottomrule
\end{tabular}
\end{sc}
\end{table*}

\subsection{Feature selection}
The small data size requires feature selection to reduce dimensionality for the initial high- dimensional feature space of 211 features. 

The selection can also convey information on the type or categories of features which capture the characteristics of the phenomena studied. This trait is very important for clinical usage and research of the disease. 

Three methods of feature selection are examined in this study: 
\subsubsection{Principal Components Analysis (PCA)}
~\\\\ PCA is applied, and the principal components which explained 95\% of the variance are chosen. 
PCA is a widely used feature selection method, and was used in previous studies on PD gait.  

However, a 
major drawback of this method in healthcare contexts is that the selected feature set contains 
linear combinations of features. The resulting transformed features can no longer be easily interpreted by clinicians.

This is especially relevant as a particular focus in the selection process is to determine the most important features in terms of discrimination between the different Parkinson's disease stages, as determined by the Hoehn and Yahr  scores.  This is not possible for the transformed PCA features. In order to address this need, two additional selection methods, that maintain the original features are implemented.

\subsubsection{Feature Selection using ANOVA}

~\\\\ The one-way ANOVA test is performed on each feature, to search for significant statistical difference, of p-value of smaller than 0.05, between the means of three or more from the different Hoehn and Yahr groups.  The features that reject the Null Hypothesis (no difference) are then selected.  
 
\subsubsection{Embedded Feature Selection: Random Forest}

~\\\\The embedded feature selection of the random forest ensemble classifier provides the features which are determined as important for the classification. The "important features" set implementation in this study consists of the features that are determined by the random forest as important in 80\% of the trials. The test of feature importance is repeated 5 times.

\subsection{Classification}
 A Random Forest (RF) classifier \cite{Breiman2001} is implemented to discriminate the different Hoehn and Yahr scores. This classifier is chosen based on two important properties for the sought discrimination: The ensemble nature of the Random Forest algorithm is well suited to the small data problem, unlike other more data hungry algorithms.
 
 Secondly, the Random Forest allows feature importance to be quantified in terms of their discrimination power. This trait is particularly relevant from the medical interpretability perspective, where the features are considered within the clinical severity classification. The classifier's hyper-parameters are optimized and tested on the full feature set and for each of the selected feature sets.  

\section{Experimental evaluation}

 The effectiveness of our method for automated discrimination of Parkinson's disease stages is 
evaluated according to two criteria, 
where the latter criterion is particularly important for clinical interpretability: 

\begin{enumerate}
    \item  Discrimination accuracy and execution time 
    \item The amount of information  gained on features and feature categories relevance to this discrimination. 
\end{enumerate}

Classification accuracy is defined as the percentage of subjects correctly classified based on the 
clinically determined Hoehn and Yahr scores.  Execution time is measured as a complementing classification efficacy metric for real-world clinical application. 

These two metrics are computed for the full feature set and for each of the three selected feature sets :  ANOVA selected features, Random Forest selected features and PCA derived features. 

Each feature set is classified by a Random Forest classifier of 100 trees, split using the Gini co-efficient. To ensure a fair comparison the number of trees is kept consistent for all selected feature sets. 

The hyper-parameter of the number of trees for the random forest is tuned experimentally for varying numbers of trees from 1-1000, where the selected number of trees provides the best trade-off of accuracy and execution time. 

Given the small data size problem, separate training, validation and test sets are not possible for evaluation. Thus, 5-fold cross-validation is performed, where the data set is randomly shuffled. 

The important feature sets selected by the ANOVA and Random Forest – the un-transformed feature sets – are then compared and evaluated according to their relevance to discrimination and clinical interpretability. 

 \begin{figure*}
 \centering
  \includegraphics[scale=0.399]{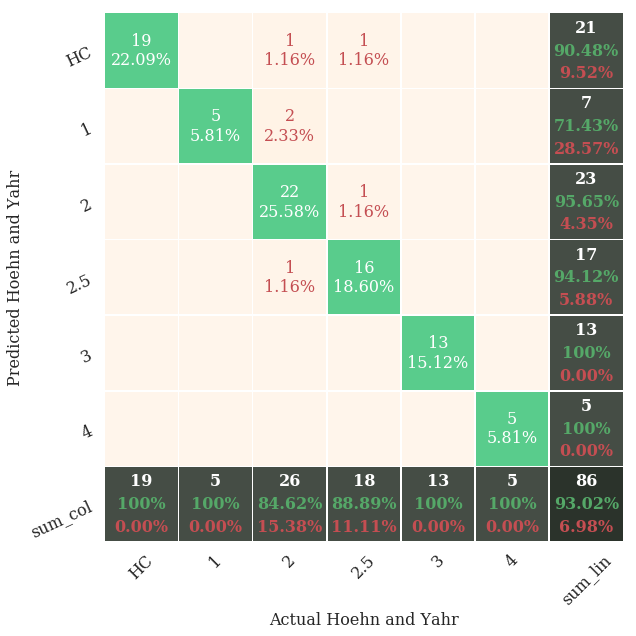}
  \caption{Confusion matrix of the Random Forest classifier using the ANOVA selected features. The horizontal axis indicates the target Hoehn and Yahr class labels: healthy controls (HC) and stages 1, 2, 2,5, 3,and 4. The vertical axis similarly indicates the output/predicted Hoehn and Yahr class labels. }
  \label{fig:conf}
\end{figure*}

\subsection{Discrimination metrics}
The performance of the Random Forest classifier in discriminating the six classes (Healthy controls and PD stages as labelled by Hoehn and Yahr scale) is presented in Table 1. Mean accuracy and its standard deviation, as well as mean execution time are presented, for the four feature sets.

The dimensionality reduction of the feature space by the selection methods was 78\% for the PCA, 68\% 
for the ANOVA and 86\% for the RF selection.  

Figure \ref{fig:conf} displays the confusion matrix for the Random Forest classifier using the ANOVA selected features.

As illustrated by the values in the diagonal of the confusion matrix in Figure 2, all healthy controls and Hoehn and Yahr 1, 3 and 4 are correctly classified. Mis-classification of 15.38\% and 11.1\% is seen for Hoehn and Yahr 2 and 2.5 respectively.
The mis-classifications of these two labels are to a Hoehn and Yahr stage lower (less severe) than the actual stage. The overall mis-classification across all class labels was 6.98\%.

\subsection{Model interpretability and clinical implications}
Only the ANOVA and RF selected features are untransformed and can provide clinical interpretability. 

\begin{figure*}[h]
 \centering
  \includegraphics[width=\linewidth]{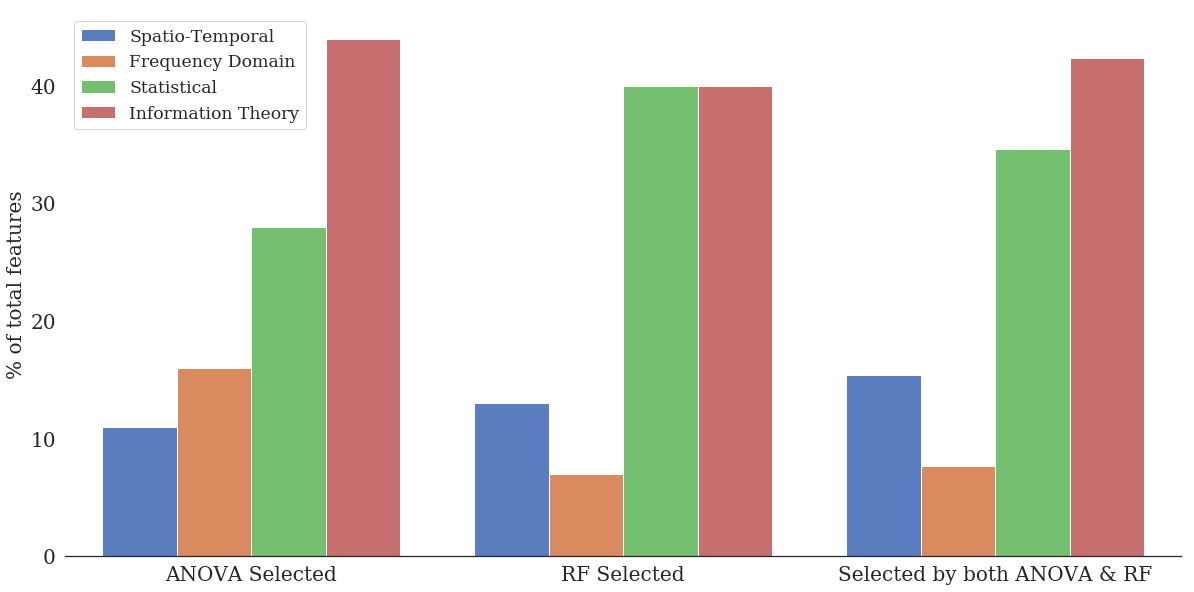}
  \caption{ Selected features grouped by feature category for the ANOVA method, RF method and those features selected by both ANOVA and RF}
  \label{feat}
\end{figure*}

Figure \ref{feat} portrays the feature categories, described in Section 3.3 which are selected by the ANOVA and RF methods, as well as the subset of features that are selected by both selection methods.  

Twenty six features are selected by both ANOVA and RF.

~\\ Their distribution according to feature categories is: 
\begin{itemize}
    \item spatio-temporal (4/26)
    \item  frequency domain (2/26)
    \item information theoretic (9/26)
    \item statistical features (11/26)
\end{itemize}

\section{Discussion}\label{discuss}
 This study presents an automated and reliable machine learning method that provides discrimination 
of Parkinson's disease stages using kinematic signals obtained from low cost walker-mounted sensors. 

The method includes signal pre-processing, feature extraction methods and an extracted feature set which extend previous studies \cite{Aharonson2018,Akbari2017,Ballesteros2017,Tahir2012,Jarchi} and  overcome the challenges introduced by low cost sensors and exo-body (walker) sensing of gait. 

The results imply that the optimal configuration for this discrimination consists of a reduced subset of ANOVA-selected features, and a random forest classifier of 100 trees. The mean accuracy achieved by this configuration is 93\%.  

The confusion matrix in Figure 2 indicates that classification errors  
typically result in mis-classification into the adjacent Parkinson's disease stage (i.e. into the Hoehn and Yahr stage immediately above or below the target stage). 

Clinical literature indicates that diagnosis errors are more frequent in stages 1 to 3, where patients are considered minimally disabled, and are still able to lead independent lives \cite{scott}. A modified Hoehn and Yahr that includes “mid-stage”  2.5 is adopted by many clinicians in an attempt  to reduce the uncertainty in the discrimination of these stages, by adding “resolution” to the scale \cite{Goetz2004}.  The scale containing “mid-scores” is, however, clinically controversial. 

In this study we use the stage 2.5 as a label and the results indicate that all mis-classifications occur between the Hoehn and Yahr stages 2 and 2.5. This may imply that the clinical label of stage 2.5 is not a “mid-score” but rather is closer to stage 2 than stage 3. The machine learning algorithm proposed in this study may then be extended to provide a finer stage division of the range 1 to 3 in the Hoehn and Yahr scale.

The comparison of three feature selection methods as presented in Table 1 reveal that all three reduced 
feature sets yielded a discrimination accuracy greater than 90\% and offered 3 to 5 times faster execution time. 

The PCA selection method provides transformed features output, which are not clinically interpretable: These feature can no longer be  individually identified and do not maintain the units of the original features. This hinders the clinical insight that can be obtained about the individual features' manifestations in patient's gait.

Between the ANOVA and RF selection methods, the ANOVA is indicated as the best trade-off. Although slower than the RF selection, it still provides a 3-times faster execution time compared to the full set, while maintaining the original set's discrimination accuracy, and has higher accuracy compared to the RF selection: 93\%, and 90\%, respectively. 

The significantly lower discrimination accuracy of the random forest-selected features may imply 
that for this data, the internal feature selection performed in the random forest was not 
able to select all of the relevant features to this multi-class discrimination, when compared to PCA 
and ANOVA.

Extending previous studies that consider mainly spatio-temporal features and/or provided binary 
discrimination of patients from controls, the present study presents a multi-class discrimination of healthy control subjects and five different PD severity stages.  The performance put emphasis on the importance of the features collected, in the context of their contribution to the discrimination. Four categories of features are evaluated by their prevalence in the machine learning feature selection subsets.   

The overlap, or common features that are selected by both ANOVA and RF, is 87\%. Interestingly, the small number of features that are selected by ANOVA and not by the RF, has contributed to a 3\% higher discrimination accuracy. 

Previous studies focused on spatio-temporal features and specifically step time, step length and step velocity.  Spatio-temporal features like step length and step velocity are among the significant features in our analysis, corroborating earlier studies. 

The current analysis extends previous studies by using signal processing and machine learning to extract a much wider variety of features, divide them into four feature categories and compare the implications of the different categories on classification performance.  

All selected subsets contain features from all four categories:  spatio-temporal-, statistical-, frequency- and information theory- based, implying that all are important for Parkinson's disease stage classification. 

The percentage of information theory and statistical features in all selected feature sets is much larger than the spatio-temporal ones. This finding implies that gait analysis of Parkinson's disease should not be limited to the observable, time-based features, but need also include these more abstract and mathematical features.  Including all four feature categories in the analysis may yield broader insight into Parkinson's disease gait deterioration with disease severity. 

The analysis of feature importance and feature category importance in our study may provide an insight for future research into gait characteristics  in Parkinson's disease. 

For example, the indication that the Hoehn and Yahr scale is based on other motor assessments in addition to gait, but could still be accurately classified using only gait features, implies a significant impact of gait impairment characteristics in the Hoehn and Yahr scale. 

The combined usability design, signal processing and machine learning discrimination proposed in this study has the potential to assist healthcare professionals in Parkinson's disease severity evaluation and facilitate a patient-centred care at a low-cost. The method is tested in a clinic, but has the potential to be used in the future by patients at home enabling quantitative PD severity assessment. This would be a step towards integration into an eHealth monitoring scheme.

{
\bibliographystyle{aaai}
\bibliography{refs, references}
}

\end{document}